\documentclass[letterpaper, 10 pt, conference]{ieeeconf}  % Comment this line out if you need a4paper

\usepackage{graphicx} % for pdf, bitmapped graphics files
\usepackage[table,xcdraw]{xcolor}
\usepackage{epsfig} % for postscript graphics files
\usepackage{amsmath} % assumes amsmath package installed
\usepackage{amssymb}  % assumes amsmath package installed
\usepackage{balance}  % balance the left-right columns
\usepackage{subcaption}
\usepackage{floatrow}
\usepackage{color,soul}
\usepackage{adjustbox}
\usepackage{verbatim} %for comments
\usepackage{url}
\usepackage[bookmarks=false]{hyperref}
\hypersetup{
     colorlinks   = true,
     citecolor    = black,
     linkcolor    = black,
     urlcolor     = black
}
\usepackage{hyphenat}

\IEEEoverridecommandlockouts                              % This command is only needed if 
\title{\LARGE \bf
Autonomous Quadrotor Landing using Deep Reinforcement Learning
}

\author{Riccardo Polvara$^{1*}$, Massimiliano Patacchiola$^{2*}$ \\ Sanjay Sharma$^{1}$, Jian Wan$^{1}$, Andrew Manning$^{1}$, Robert Sutton$^{1}$ and Angelo Cangelosi$^{2}$ % <-this % stops a space
\thanks{*Both first and second author contributed equally and should be considered co-first authors.}% <-this % stops a space
\thanks{$^{1}$Autonomous Marine System Research Group, School of Engineering, Plymouth University, UK
        {\tt\small [corresponding author] riccardo.polvara@plymouth.ac.uk}}%
\thanks{$^{2}$Centre for Robotics and Neural Systems, School of Computing, Electronics and Mathematics, Plymouth University, UK
    }%
}

\begin{document}

\maketitle
\begin{abstract}
Landing an unmanned aerial vehicle (UAV) on a ground marker is an open problem despite the effort of the research community. Previous attempts mostly focused on the analysis of hand-crafted geometric features and the use of external sensors in order to allow the vehicle to approach the land-pad. In this article, we propose a method based on deep reinforcement learning that only requires low-resolution images taken from a down-looking camera in order to identify the position of the marker and land the UAV on it. The proposed approach is based on a hierarchy of Deep Q-Networks (DQNs) used as high-level control policy for the navigation toward the marker. We implemented different technical solutions, such as the combination of vanilla and double DQNs, and a partitioned buffer replay. Using domain randomization we trained the vehicle on uniform textures and we tested it on a large variety of simulated and real-world environments. The overall performance is comparable with a state-of-the-art algorithm and human pilots.

%Learning is achieved without any human supervision, giving to the agent an high-level feedback.
 %\textcolor{red}{The results show that policies trained on simple uniform textures can autonomously accomplish landing on a large variety of unknown simulated environments of increasing complexity. The overall performance is comparable with a state-of-the-art algorithm and human pilots.}
%The results show that the quadrotor can autonomously accomplish landing on a large variety of simulated environments and with relevant noise. In some conditions the DQN outperformed human pilots tested in the same environment. 
\end{abstract}

\section{INTRODUCTION}
In the upcoming years an increasing number of autonomous systems will pervade urban and domestic environments. 
The next generation of Unmanned Aerial Vehicles (UAVs) requires high-level controllers in order to move in unstructured environments and perform multiple tasks. Recently a new application has been proposed, namely the use of quadrotors for the delivery of packages and goods. In this scenario the most delicate part is the identification of a ground marker and the vertical descent maneuver. Previous works used hand-crafted features analysis and external sensors in order to identify the land-pad. In this work we propose a completely different approach, based on recent breakthroughs achieved with Deep Reinforcement Learning (DRL) \cite{mnih2015human}. Our method only requires low-resolution images acquired from a down-looking camera that are given as input to a hierarchy of Deep Q-Networks (DQNs). The output of the networks is a high level command that directs the drone toward the marker. The most remarkable advantage of DRL is the absence of any human supervision, allowing the quadrotor to autonomously learn how to use high-level actions in order to land.

\begin{figure}[t!]
    \centering
    \includegraphics[width=0.92\textwidth]{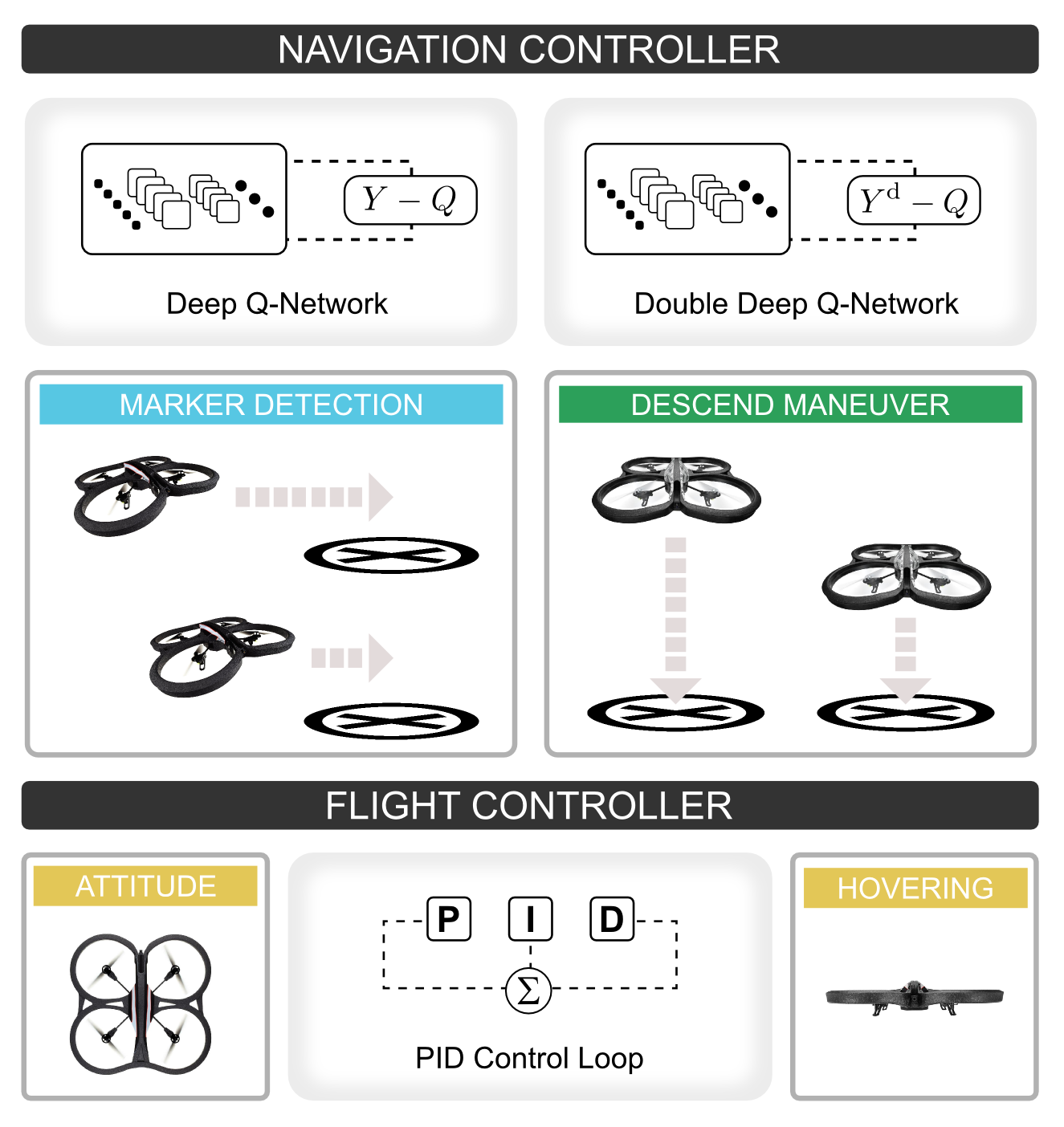}
    \caption{System overview. The navigation controller is built on top of the flight controller. The marker detection and the descent maneuver are achieved through two distinct DQNs.}
    \label{fig:fig_system_overview}
\end{figure}

The use of DRL in the landing problem is not straightforward. Previous applications mainly focused on deterministic environments such as the Atari game suite \cite{mnih2015human}. Using DRL in unstructured environments with robotic platforms has had limited success.
In this work we tackled the landing problem introducing different technical solutions. We used a divide-and-conquer strategy and we split the problem in two sub-tasks: landmark detection and vertical descent. Two specialized DQNs take care of the two tasks and are connected through an internal trigger engaged by the networks itself. Moreover, we used \textit{double DQN} \cite{van2016deep} to reduce overestimation problems that commonly arise when the agent moves in complex environments. To solve the issue of sparse and delayed reward we implemented a new type of prioritized experience replay, called partitioned buffer replay, that splits the experiences in multiple containers and guarantees the presence of rare transitions in the training batch. As far as we know, the present work is the first to use an unsupervised learning approach to tackle the landing problem. We show an overview of the system in Figure~\ref{fig:fig_system_overview} and a video in our repository \footnote{https://github.com/pulver22/QLAB/tree/master/share/video}.
%In Figure~\ref{fig:fig_system_overview} we show an overview of the system.\footnote{https://github.com/pulver22/QLAB/tree/master/share/video}

%We shortly summarize the contribution of our work in three points. (i) As far as we know the present work is the first to use an unsupervised learning approach to tackle the landing problem. We trained the agent using only low-resolution images, without the need of direct supervision and hand-crafted features. This method represents a significant improvement compared to previous work. (ii) We introduce new technical solutions such as a hierarchy of DQNs which are able to autonomously trigger each other, and a new form of prioritized buffer replay. Those techniques are not limited to the landing problem and can be used in other complex tasks. (iii) Using the proposed method we trained a commercial UAV in multiple simulated environments. The results on a test benchmark showed performances which are similar to human pilots tested in the same conditions.

%We shortly summarize the contribution of our work in three points. (i) As far as we know the present work is the first to use an unsupervised learning approach to tackle the landing problem. (ii) We introduce new technical solutions such as a hierarchy of DQNs, and a prioritized buffer replay. (iii) Using the proposed method we trained a commercial UAV in multiple simulated environments obtaining performances that are similar to human pilots and a state-of-the-art algorithm.

\section{Related work}\label{sec:state_of_the_art}

In this section we present a brief literature review in order to offer an overview on the topic. This review is not meant to be complete, and it only aims to show how our method differentiates from previous work. 

We can broadly group in three classes the methods used for landing UAVs: sensor-fusion, device-assisted, and vision-based. The sensor-fusion methods rely on the use of multiple sensors, in order to gather enough data for a robust pose estimation. In a recent work \cite{forster2015continuous} the data from a downward-looking camera and an inertial measurement unit were combined in order to build a three-dimensional reconstruction of the terrain. Given the two-dimensional elevation map was possible to find a secure surface area for landing.
In \cite{saripalli2006landing} the authors used a particular geometric shape for the landing pad in conjunction with analysis of multiple sensors in order to accurately estimate the position of the drone with respect to the marker.
A ground-based multisensor fusion system has been proposed in \cite{zhou2015autonomous}. The system included a pan-tilt unit, an infrared camera and an ultra-wideband radar used to center the UAV in a recovery area and guide it toward the ground. A similar work is presented in \cite{polvara2017toward} in order to land on an AR-tag marker posed on a moving vessel.
%An extended Kalman filter was then used for estimating the trajectory and guiding the drone toward the marker.

%Device-assisted
Device-assisted methods rely on the use of ground sensors in order to precisely estimate the position and trajectory of the drone.
A system based on infra-red lights has been used in \cite{gui2013airborne}. The authors adopted a series of parallel infrared lamps disposed in a runway. The camera on the vehicle was equipped with optical filters for capturing the infrared lights and the images were forwarded to a control system for pose estimation. A Chan-Vese approach supplemented through an extended Kalman filter has been proposed in \cite{tang2016ground} for ground stereo-vision detection.

%The landing of a fixed-wing UAV has been accomplished in \cite{kong2017localization} using a ground-based vision system. Two cameras, mounted on two separate pan-tilt units, were placed on the sides of an airport runway and remotely controlled to detect the position of the UAV.

%Vision-Based

The vision based approaches analyse geometric features in order to find ground pads and land.
A method based only on a monocular camera has been proposed in \cite{lange2009vision}. The system used a well defined target pattern, easy to identify at different distances. Having a series of concentric circles, it was proved to be possible to find the landmark also when partially occluded.
A modified version of the international landing pattern has been used in \cite{lin2017monocular}. The solution adopted used a seven-stages vision algorithm to identify and track the pattern in a cluttered environment and reconstruct it when partially observable.
The use of AR-tag fiducial marker has been taken into account in \cite{falanga2017vision} and \cite{vetrella2018improved}. In both cases a precise pose estimation has been done using only an onboard camera.
In \cite{lee2012autonomous} a vision-based visual servoing algorithm has been used to track a moving platform and to produce velocity commands for an adaptive sliding controller.

%A recent work \cite{falanga2017vision} used computer vision algorithm for detecting a moving target using only an onboard camera. The information was used to precisely estimate the quadrotor pose. \textcolor{red}{A similar work is present in \cite{vetrella2018improved}, and is based on tunable spatio-temporal trajectories that are generated using a bio-guidance strategy. Both the previous works made use of AR-tag fiducial markers to identify the landing spot.}

The previous works showed different limitations that we discuss here.
Sensor-fusion methods often use information gathered from expensive sensors that cannot be integrated in low-cost drones. Most of the time these methods rely on the contribution of GPS that may be unavailable in real-world scenarios.
The device-assisted approaches allow obtaining an accurate estimation of the drone pose. However the use of external devices is not always possible because they are not always available. Vision-based methods have the advantage of using only on-board sensors and mainly rely on cameras. The main limit of these methods is that low-level features are often viewpoint-dependent and subject to failure in ambiguous cases. The present work directly deals with all the aforementioned problems. Our solution is based only on a monocular onboard camera and does not use any other sensors or external devices. The use of DQNs significantly improves the marker detection and is robust to projective transformations and marker corruption. 
%The method is fully unsupervised and does not require any hand-crafted geometric feature.

\section{Proposed method}
In this section we describe the landing problem in reinforcement learning terms and we present the technical solutions we adopted.

\subsection{Problem definition and notation}\label{ssec:problem_definition_and_notation}
Here we consider the landing problem as divided in two sub-problems: landmark detection and vertical descent. The detection requires an exploration on the xy-plane, where the quadrotor has to horizontally shift in order to align its body frame with the marker. In the vertical descent phase the vehicle has to reduce the distance from the marker using vertical movements. Moreover, the drone has to shift on the xy-plane in order to keep the marker centered. 

%A graphical representation of the two phases is reported in Figure~\ref{fig:fig_example_landmark_detection_vertical_descent}.

%\begin{figure}[b]
%   \begin{subfigure}[b]{0.49\textwidth}
%   \centering
%   \includegraphics[width=\linewidth]{images/fig_example_landmark_detection.png}
%   \caption{}
%   \label{fig:fig_example_landmark_detection_vertical_descent_a} 
%\end{subfigure}
%\hfill
%\begin{subfigure}[b]{0.49\textwidth}
%   \includegraphics[width=\linewidth]{images/fig_example_vertical_descent.png}
%   \caption{}
%   \label{fig:fig_example_landmark_detection_vertical_descent_b}
%\end{subfigure}
%\caption{Representation of landmark detection (a) and vertical descent (b). }
%\label{fig:fig_example_landmark_detection_vertical_descent}
%\end{figure}

\begin{figure}[t!]
    \centering
    \includegraphics[width=0.95\textwidth]{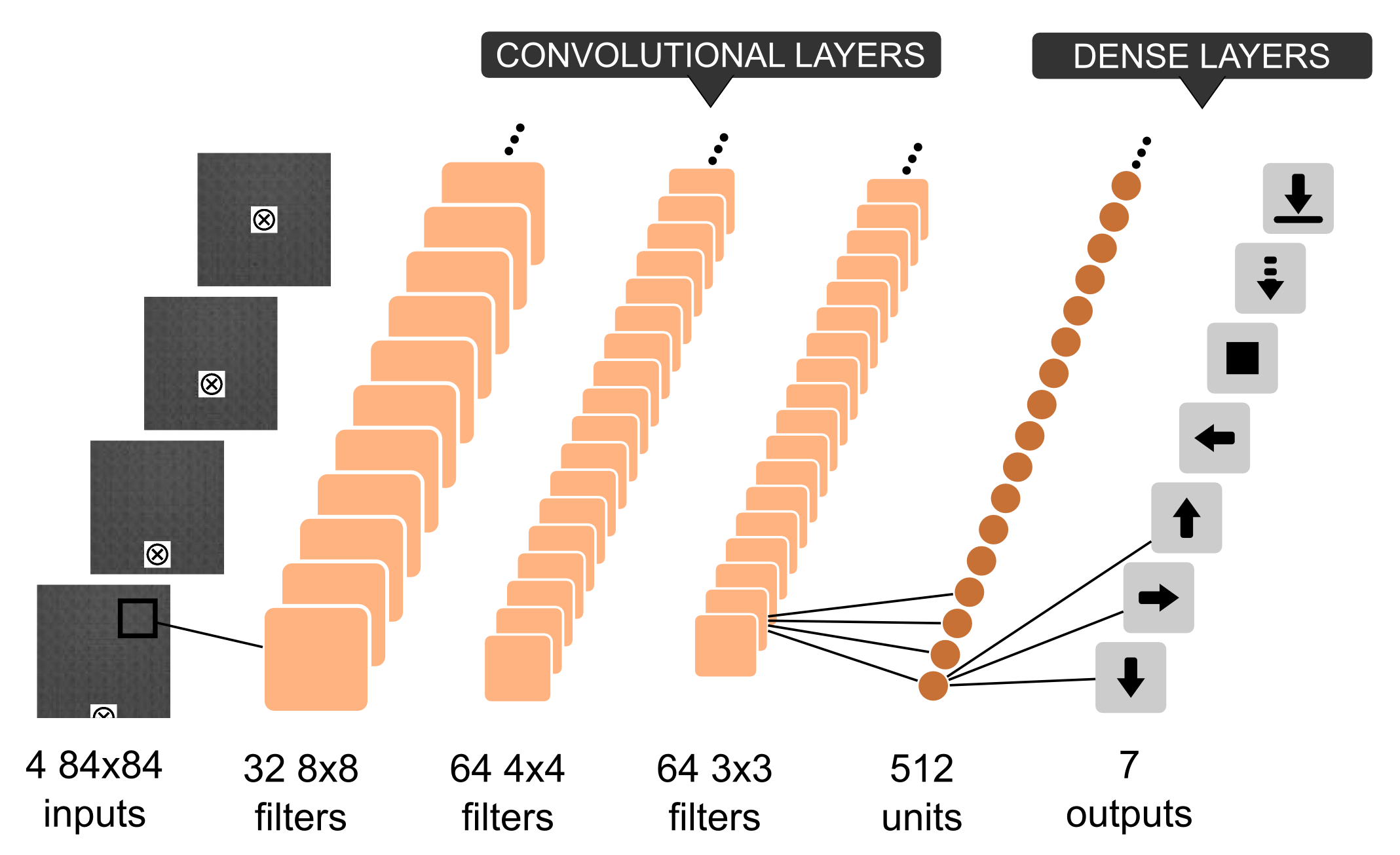}
    \caption{Graphical representation of the DQN. The network takes in input four $84\times84$ images, and generates in output 7 actions: forward, right, backward, left, stop, descent, trigger.}
    \label{fig:fig_convolutional_neural_network_architecture}
\end{figure}

Formally both the problems can be reduced to Markov Decision Processes (MDPs). At each time step $t$ the agent receives the state $s_{t}$, performs an action $a_{t}$ sampled from the action space $A$, and receives a reward $r_{t}$ given by a reward function $R(s_{t}, a_{t})$. The action brings the agent to a new state $s_{t+1}$ in accordance with the environmental transition model $T(s_{t+1} | s_{t}, a_{t})$. In the particular case faced here the transition model is not given (model free). The goal of the agent is to maximize the discounted cumulative reward called return $R=\sum_{k=0}^{\infty} \gamma^{k} r_{t+1}$, where $\gamma$ is the discount factor. Given the current state the agent can select an action from the internal policy $\pi = P(a | s)$. In off-policy learning the prediction of the cumulative reward can be obtained through an action-value function $Q^{\pi}(s,a)$ adjusted during the learning phase in order to approximate $Q^{*}(s,a)$, the optimal action-value function. In this work we use a Convolutional Neural Network (CNN) for approximating the Q-function following the approach presented in \cite{mnih2015human}. 
%Recently, CNNs achieved outstanding results in a large variety of problems, such as image classification \cite{krizhevsky2012imagenet} and head pose estimation \cite{patacchiola2017head}. 
The CNN takes as input four $84 \times 84$ grey scale images acquired by the downward looking camera mounted on the drone. The images are processed by three convolutional layers and two fully connected layers. As activation function we used the rectified linear unit. The first convolution has 32 kernels of $8 \times 8$ with stride of 2, the second layer has 64 kernels of $4 \times 4$ with strides of 2, the third layer convolves 64 kernels of $3 \times 3$ with stride 1. The fourth layer is a fully connected layer of 512 units followed by the output layer that has a unit for each valid action (backward, right, forward, left, stop, descent, land). Depending on the simulation, we used a sub-set of the total actions available, we refer the reader to Section \ref{sec:experiments} for additional details. A graphical representation of the network is presented in Figure~\ref{fig:fig_convolutional_neural_network_architecture}. 

It is important to focus on the two phases that characterize the landing problem in order to isolate important issues.
In the landmark detection phase we made the reasonable assumption of a flight at fixed-altitude. The vertical alignment with the landmark is obtained through shifts in the xy-plane. This expedient does not have any impact at the operational level but dramatically simplifies the task. To adjust $\theta$, the parameters of the DQN, we used the following loss function:

\begin{equation}\label{eq:eq_loss_function}
    L_{i}(\theta_{i}) = \mathop{\mathbb{E}_{(s,a,r,s^{\prime}) \sim U(D)}} \bigg[ \big(  Y_{i} - Q(s,a; \theta_{i}) \big)^{2} \bigg]
\end{equation}

with $D = {(e_{1}, ... , e_{t})}$ being a dataset of experiences $e_{t} = (s_{t}, a_{t}, r_{t}, s_{t+1})$ used to uniformly sample a batch at each iteration $i$. The network $Q(s,a; \theta_{i})$ is used to estimate actions at runtime, whereas $Y_{i}$ is the target that is defined as follows:

\begin{equation}\label{eq:eq_target}
    Y_{i} =  r + \gamma \max_{a^{\prime}} Q(s^{\prime}, a^{\prime}; \theta_{i}^{-})
\end{equation}

the network $Q(s^{\prime}, a^{\prime}; \theta^{-}_{i})$ is used to generate the target and is constantly updated. The use of the target network is a trick that improves the stability of the method. The parameters $\theta$ are updated every $C$ steps and synchronized with $\theta^{-}$. In the standard approach the experiences in the dataset $D$ are collected in a preliminary phase using a random policy. The dataset $D$ is also called buffer replay and it is a way to randomize the samples breaking the correlation and reducing the variance \cite{wawrzynski2013autonomous}.

The vertical descent phase is a form of Blind Cliffwalk \cite{schaul2015prioritized} where the agent has to take the right action in order to progress through a sequence of $N$ states and finally get a positive or a negative reward. The intrinsic structure of the problem makes extremely difficult to obtain a positive reward because the target-zone is only a small portion of the state space. The consequence is that the buffer replay does not contain enough positive experiences, making the policy unstable. To solve this issue we used a form of buffer replay, called partitioned buffer replay, that discriminates between rewards and guarantees a fair sampling between positive, negative and neutral experiences. Another issue connected with the reward sparsity is the utility overestimation \cite{thrun1993issues}. During a preliminary research we observed this problem in the vertical descent phase. Monitoring the Q-max value (the highest utility returned by the Q-network) we noticed that it rapidly increased, overshooting the maximum possible utility of 1.0. The overestimation was associated with all the actions but the trigger. The trigger leads to a terminal state, therefore its utility is updated without the $max$ operator. The $max$ operator has been found to be the responsible of the overestimation in deep Q-learning \cite{van2016deep}. In our case the overestimated utilities of the four horizontal movements (grown up to 2.0 after $10^{5}$ frames) were higher than the non-overestimated utility associated with the trigger (stably converged to 1.0). As a result the drone moved on top of the marker and then shifted on the xy-plane without engaging the trigger. A solution to overestimation has been recently proposed and has been called double DQN \cite{van2016deep}. The target estimated through double DQN is defined as follows:

\begin{equation}\label{eq:eq_target_double}
 Y_{i}^{\text{d}} =  r + \gamma \ Q(s^{\prime}, \underset{a^{\prime}}{\mathrm{argmax}} \ Q(s^{\prime}, a^{\prime}; \theta_{i}); \theta_{i}^{-})
\end{equation}

Using this target instead of the one in Equation \ref{eq:eq_target} the divergence of the DQN action distribution is mitigated resulting in a faster convergence and increased stability.

\subsection{Partitioned buffer replay} \label{ssec:partitioned_buffer_replay}
%In MDPs with a sparse and delayed reward it is difficult to obtain a positive feedback. In these cases the experiences accumulated in the buffer replay may be extremely unbalanced. Neutral transitions are frequent and for this reason they are sampled with an high probability, whereas positive and negative experiences are uncommon and difficult to sample. 
%The landing problem is a form of Blind Cliffwalk \cite{schaul2015prioritized}. The drone has to move vertically and horizontally in order to reach the ground pad. The positive reward is extremely sparse and delayed because it is obtained on a small portion of the state space.
In a preliminary research we find out that the vertical descent was affected by the sparsity of positive and negative rewards. The shortage of positive and negative experiences caused an underestimation of the utilities associated to the triggers.
To deal with sparse rewards it has been proposed to divide the experiences in two buckets, one with high priority and the other with low priority \cite{narasimhan2015language}. Our approach is an extension of this method to $K$ buckets. Another form of prioritized buffer replay has been proposed in \cite{schaul2015prioritized}. The authors suggest to sample important transitions more frequently. The prioritized replay estimates a weight for each experience based on the temporal difference error. Experiences are sampled with a probability proportional to the weight. The limitation of this form of prioritization is the introduction of another layer of complexity that may not be justified for applications were there is a clear distinction between positive and negative rewards. Moreover this method requires $O(\log N)$ to update the priorities. This issue does not significantly affect performances on the standard benchmark but it has a relevant effect on robotics application, where there is a high cost in obtaining experiences.

In Section \ref{ssec:problem_definition_and_notation} we defined  $D^{} = {(e_{1}, ... , e_{t})}$ being a dataset of experiences $e = (s, a, r, s^{\prime})$ used to uniformly sample a batch at each iteration $i$. To create a partitioned buffer replay we have to divide the reward space in $K$ partitions:

\begin{equation}
R = R(s,a) \rightarrow \text{Im} \ R \ = R_{1} \cup ... \cup R_{K}
\end{equation}

For any experience $e_{i}$ we associate its reward $r_{i} = r(e_{i})$ and we define the $K$th buffer replay:

\begin{equation}
D_{K} = \{ (e_{1}, ..., e_{N}) : r_{1}, .., r_{N} \in R_{K} \}
\end{equation}

The batch used for training the policy is assembled picking experiences from each one of the $K$ datasets with a certain fraction $\rho \in \{\rho_{1}, ..., \rho_{K} \}$.

In our particular case we have $K=3$, meaning that we have three datasets with $D^{+}$ containing experiences having positive rewards, $D^{-}$ containing experiences having negative rewards, and $D^{\sim}$ for experiences having neutral rewards. The fraction of experiences associated to each one of the dataset is defined as $\rho^{+}$, $\rho^{-}$, and $\rho^{\sim}$.

When using a partitioned buffer replay there is a substantial increase in the available number of positive and negative experiences. For instance using a single buffer of size $2 \times 10^{4}$ and accumulating $8.4 \times 10^{4}$ transitions, the total number of positive experiences is 343 and the number of negative experiences is 2191. Using a partitioned buffer with size $2 \times 10^{4}$ for the neutral partition, and size $10^{4}$ for positive and negative partitions, the total number of positive experiences is 1352 and the number of negative experiences 9270.

\subsection{Hierarchy of DQNs}

\begin{figure}
    \centering
    \includegraphics[width=\textwidth]{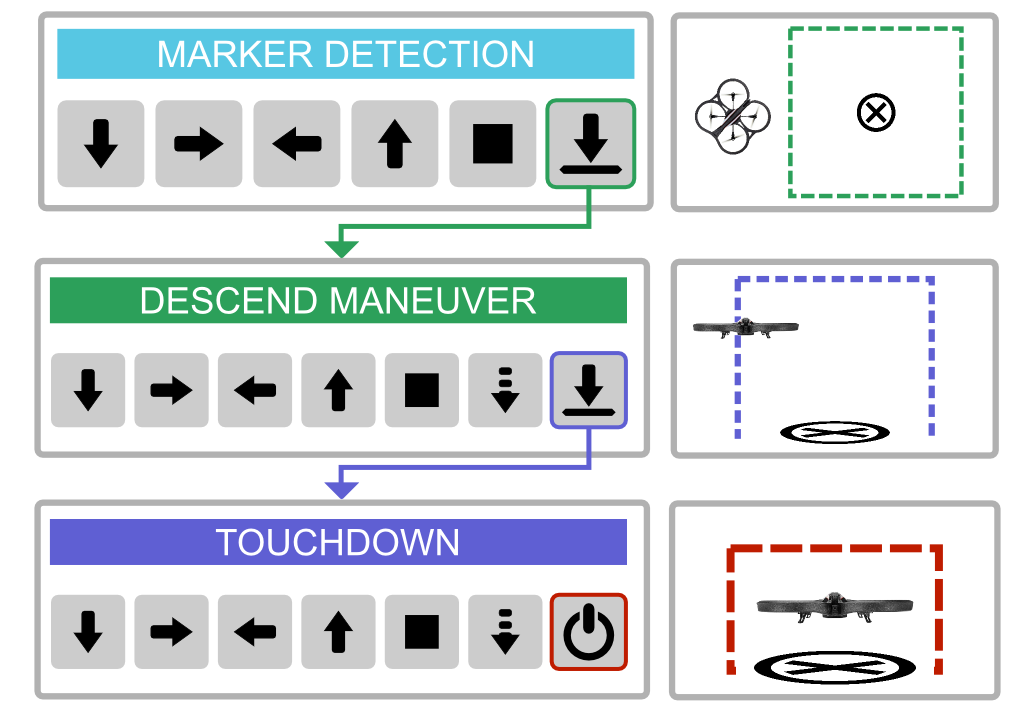}
    \caption{Finite-state machine for autonomous landing based on the DQN hierarchy method. Each state has a specific trigger that enables the DQN in the next stage.}
    \label{fig:fig_state_machine.png}
\end{figure}

Our method is based on the use of a hierarchy of DQNs representing sub-policies used to deal with different phases of the navigation.
Similarly to a finite-state machine the global policy is divided into modules and each module is governed by a specific DQN or control loop. The DQNs are able to autonomously understand when it is time to call the next state. The advantages of such a method are twofold. On the one hand it is possible to reduce the complexity of the task using a divide-and-conquer approach. On the other hand, the use of a function approximator is confined in specific sandboxes making their use in robotic applications safer. 

A similar approach is described in hierarchical reinforcement learning \cite{barto2003recent} where a set of sub-policies, called options, are available to the agent in specific states. The options control the agent in sub-regions of a core MDP called semi-MDPs. In the present work we assume that the core MDP can be divided into multiple isolated instances and that each instance is a proper MDP. The advantage is that we can use standard Q-learning to train the agent.

%Moreover we assume that it is possible to combine the auxiliary MDPs in a ordered sequence and connect the elements of the sequence through shared states. In shared states the use of particular actions, called triggers, enables the passage to the next MDP. The triggers are additional actions that do not belong to the action space of the core MDP. Engaging the triggers in shared states leads to a reward that is equal to the maximal reward associated with the core MDP. How to define the auxiliary MDPs is an operation left to the designer that requires some knowledge of the core MDP. However we do not exclude that such an operation can be automatized using some form of cluster analysis.}

The finite MDP describing the landing problem can be divided in three main stages: landmark detection, descent maneuver, touchdown. We described in Section \ref{ssec:problem_definition_and_notation} the first two phases. The touchdown consists in decreasing the power of the motors in the last few centimeters of the descent and then safely deactivate the UAV components (e.g. motors, cameras, boards, control unit, etc.). In this article we mainly focused on the first two stages, because they represent the most challenging part of the landing procedure.
A graphical representation of a hierarchical state machine is represented in Figure~\ref{fig:fig_state_machine.png}.
We trained the first DQN (marker detection) to receive a positive reward when the trigger was enabled inside a target area. Negative reward was given if the trigger was enabled outside the target area. The second network (descent maneuver) was trained using the same idea. In a preliminary phase we also trained a single network to achieve both detection and descending. Given the size of the combined spaces the network was not able to converge to a stable policy. As a baseline we also report the accumulated reward curve of this network in Section \ref{sec:experiments}.

%Once the networks have been trained it is possible to assemble the state-machine pipeline. 

\begin{figure}
    \centering
    \includegraphics[width=\textwidth]{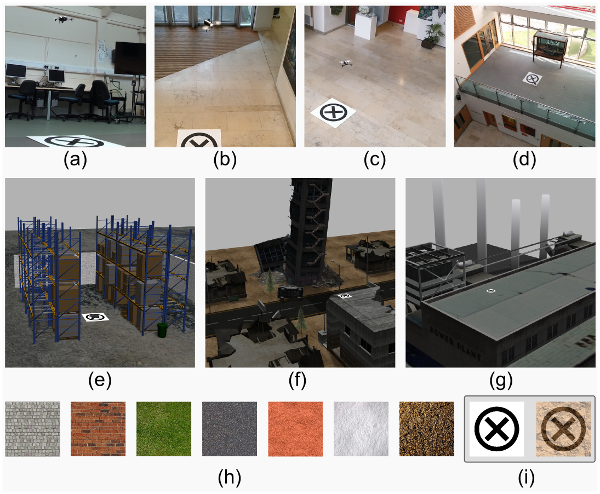}
    \caption{Real environments: laboratory (a), small hall (b), large hall (c), mezzanine (d). Photo-realistic environments: warehouse (e), disaster site (f), powerplant (g). Textures (h): pavement, brick, grass, asphalt, sand, snow, soil. Marker and corrupted marker (i).}
    \label{fig:fig_simulator_textures_sample}
\end{figure}

\subsection{Training through domain randomization}
The reality gap is the obstacles that makes it difficult to implement many robotic solutions in real world. This is especially true for DRL where a large number of episodes is necessary in order to obtain stable policies. Recent research worked on bridging this gap using domain transfer techniques. An example is domain randomization \cite{tobin2017domain}, a method for training models on simulated images that transfer to real images by randomizing rendering in the simulator. Here we adopt domain randomization in order to train the UAV in simple simulated environments and test it in complex environments (both simulated and real). The remarkable property of this approach is that it does not require any pre-training on real images. If the variability is significant enough, models trained in simulation generalize to the real world with no additional training. In the next session we show how domain randomization has been included in the training phase and how the experiments have been organized.

\section{Experiments}\label{sec:experiments}
In Section \ref{ssec:first_simulation} the methodology and the results obtained with the DQN specialized in the landmark detection phase is presented, while in Section \ref{ssec:second_simulation} those concerning the vertical descent phase. In both training and testing we used the same environment (Gazebo 7.7.x, ROS Kinetic) and drone (Parrot BeBop 2). The simulator is a fork of the one used in \cite{engel2012camera} and it is freely available on our repository\footnote{https://github.com/pulver22/QLAB}. The control command sent to the vehicle is represented by a continuous vector $\in [-1,1]$ that allows moving the drone with a specific velocity on the three axes.
We must point out that the physics of the engine has not been simplified in any way. There are important oscillatory effects during accelerations and decelerations that introduces a swinging behaviour with consequent perspective distortion in the images acquired. Moreover a summation of forces effect shows when the vehicle accumulates inertia and a new velocity command is given. The DRL algorithm has to deal with this source of noise.  

\subsection{First series of simulations} \label{ssec:first_simulation}
In the first series of simulations we trained and tested the DQNs for the marker detection phase. We considered two networks having the same structure (Figure~\ref{fig:fig_convolutional_neural_network_architecture}) and we trained them in two different conditions. The first network was trained with a uniform asphalt texture (DQN-single), whereas the second network was trained with multiple textures (DQN-multi). The ability to generalize to new unseen situations is very important and it should be seriously taken into account in the landing problem. Training the first network on a single texture is a way to quantify the effect of a limited dataset on the performance of the agent. In the DQN-multi condition the networks were trained using seven different groups of textures: asphalt, brick, grass, pavement, sand, snow, soil (Figure~\ref{fig:fig_simulator_textures_sample}-h). These networks should outperform the ones trained in the condition with single texture.

At each episode the drone started at a fixed altitude of 20 m that was maintained for the entire flight. This expedient was useful for two reasons: it significantly reduced the state space to explore and it allowed visualizing the marker in most of the cases giving a reference point for the navigation. In a practical scenario this solution does not have any impact on the flight, the drone is kept at a stable altitude and the frames are acquired regularly. 
To stabilize the flight we introduced discrete movements, meaning that each action was repeated for 2 seconds and then stopped leading to an approximate shift of 1 meter, similarly to the no-operation parameter used in \cite{mnih2015human}. The frames from the camera were acquired between the actions ($0.5$ Hz) when the vehicle was stationary. This expedient stabilized convergence reducing perspective errors.

\subsubsection{Methods}\label{ssec:first_simulation_methods}
\begin{figure}
   \begin{subfigure}[b]{0.48\textwidth}
   \centering
   \includegraphics[width=\linewidth]{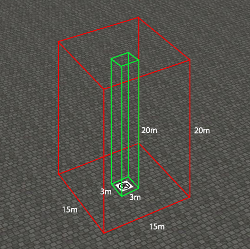}
   \caption{}
   \label{fig:fig_simulator_bounding_boxes_a} 
\end{subfigure}
\hfill
\begin{subfigure}[b]{0.48\textwidth}
   \includegraphics[width=\linewidth]{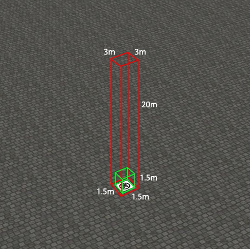}
   \caption{}
   \label{fig:fig_simulator_bounding_boxes_b}
\end{subfigure}
\caption{Flying-zone (red) and target-zone (green) for landmark detection (a) and vertical descent (b).}
\label{fig:fig_simulator_bounding_boxes}
\end{figure}

The training environment was represented by a uniform texture of size $100 \times 100$ m with the landmark positioned in the center. The environment contained two bounding boxes (Figure \ref{fig:fig_simulator_bounding_boxes_a}). At the beginning of each episode the drone was spawned at 20 m of altitude inside the perimeter of the larger bounding box ($15\times15\times20$ m) with a random position and orientation. A positive reward of 1.0 was given when the drone activated the trigger in the target-zone, and a negative reward of -1.0 was given if the drone activated the trigger outside the target-zone. A negative cost of living of -0.01 was applied to all the other conditions. A time limit of 40 seconds (20 steps) was used to stop the episode and start a new one. In the DQN-multi condition the ground texture was changed every 50 episodes and randomly sampled between the 71 available. The target and policy networks were synchronized every $C=10000$ frames. The agent had five possible actions available: forward, backward, left, right, land-trigger. The action was repeated for 2 seconds, then the drone was stopped and a new action was sampled. The buffer replay was filled before the training with $4 \times 10^{5}$ frames using a random policy. We trained the two DQNs for $6.5 \times 10^{5}$ frames. We used an $\epsilon$-greedy policy with $\epsilon$ decayed linearly from 1.0 to 0.1 over the first $5 \times 10^{5}$ frames and fixed at 0.1 thereafter. The discount factor $\gamma$ was set to 0.99. As optimizer we used the RMSProp algorithm with a batch size of 32. The weights were initialized using the Xavier initialization method. The DQN algorithm was implemented in Python using the Tensorflow library. Simulations were performed on a workstation with an Intel i7 (8 core) processor, 32 GB of RAM, and the NVIDIA Quadro K2200 as graphical processing unit. On this hardware the training took 5.2 days to complete.

%To test the performance of the policies we measured the landing success rate of both DQN-single and DQN-multi in a new environment using 21 unknown textures. 
To test the performance of the policies we measured the detection success rate of both DQN-single and DQN-multi in six tests. (i) The first test was performed on 21 unknown uniform textures belonging to the same categories of the training set. (ii) The second test was done on the same environments but at different altitudes (20, 15, and 10 meters). (iii) The third test was performed on the same 21 unknown textures but using a marker corrupted through a semi-transparent dust-like layer. (iv) The fourth test was done randomly sampling 25 textures from the test set and mixing them in a mosaic-like composition. (v) The fifth test has been done on three photo-realistic environments namely a warehouse, a disaster site, and a power-plant (Figure \ref{fig:fig_simulator_textures_sample}-e/g). (vi) The sixth and last test consisted in a real-world implementation in the mezzanine environment (Figure \ref{fig:fig_simulator_textures_sample}-d). The mezzanine is the only environment that allowed flying at an high altitude. We also measured the performances of a random agent, an AR-tracker algorithm \cite{polvara2017toward}, and human pilots in all the simulated environments. The human data has been collected using two methodologies. In the first approach 7 volunteers used a space-navigator mouse that gave the possibility to move the drone in the three dimensions at a maximum speed of 0.5 m/s. In the second methodology 5 volunteers used a keyboard to move the drone in four directions on the xy-plane through discrete steps of 1 meter. The first methodology has been adopted in order to give to the subjects a natural control interface, whereas the second methodology gave the same control conditions of the drone.
In both conditions preliminary training allowed the subject to familiarize itself with the task. After the familiarization phase the real test started. In the landmark detection the subjects had to align the drone with the ground marker and trigger the landing procedure when inside the target-zone. The subjects performed five trials for each one of the environments contained in the test set (randomly sampled). A time limit of 40 seconds (20 steps) was applied to each episode. A landing attempt was declared as failed when the time limit expired or when the subject engaged the trigger outside the target-zone.

\subsubsection{Results}
\begin{figure}
    \centering
    \includegraphics[width=\textwidth]{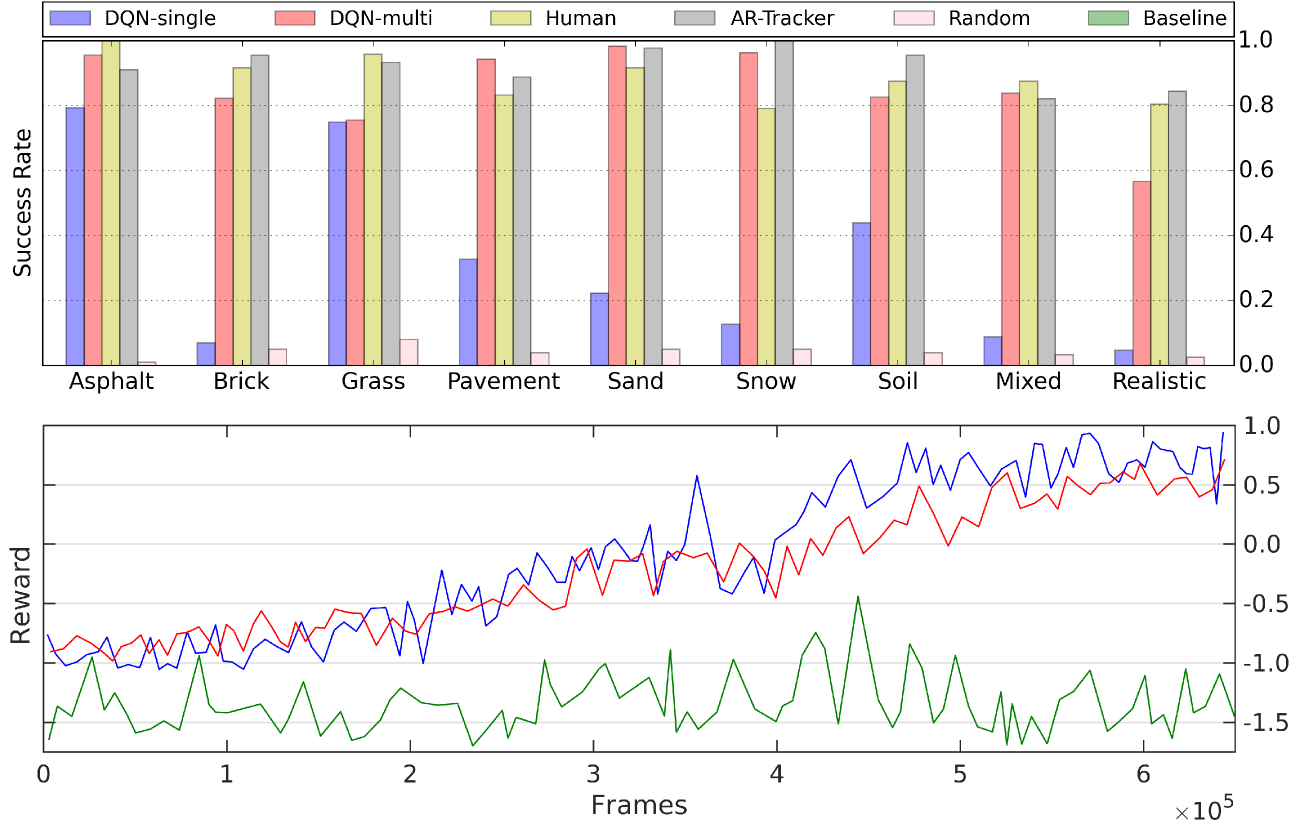}
    \caption{Results of the first series of simulations. Top: detection success rate. Bottom: accumulated reward per episode for DQN-single (blue line), DQN-multi (red-line), and baseline (green-line).}
    \label{fig:fig_results_first_series}
\end{figure}

% DQN-single and DQN-multi standard
The results for both DQN-single and DQN-multi show that the agents were able to learn an efficient policy for maximizing the reward. In both conditions the reward increased stably without any anomaly (Figure~\ref{fig:fig_results_first_series} bottom). In the same figure we also report the reward curve for a baseline condition, where a single network has been trained to perform both detection and descending. The reward of the baseline did not increase significantly and the resulting policy was unable to engage the trigger inside the target-zone. The results of the test phase are summarized in Figure~\ref{fig:fig_results_first_series} (top). The bar chart compares the performances of DQN-single, DQN-multi, human pilots, AR-tracker and random agent. For human pilots we only report the results for the discrete control condition, since the score was higher than the space-navigator condition ($+6\%$). The average score on the first test (uniform textures) for the DQN-multi is $91\%$. The score obtained by the agent trained on a single texture (DQN-single) are significantly lower ($39\%$). The human performance is $90\%$, whereas the AR-tracker has an average score of $95\%$. The random agent has an average reward of $4\%$ in this environment. Since both human pilots and DQNs used discrete steps to move in the environments, it is possible to estimate the average number of discrete steps required to accomplish detection. For human pilots the average number of steps is 12, whereas for the DQN-multi is 6, meaning that humans were significantly slower. Testing the DQN-multi at different altitudes we noticed that the accuracy increased at 15 ($95\%$) and 10 ($93\%$) meters, with respect to the accuracy at the training altitude of 20 meters ($89\%$). This result is explained by the fact that at lower altitudes the marker is more visible. In the third test we compared the DQN-multi and AR-tracker on uniform textures using the corrupted marker. We observed a significant drop in the AR-tracker performances from $94\%$ to $0\%$ explained by the fact that the underlying template matching algorithm failed in identifying the corrupted marker. In the same condition the DQN-multi performed well, with a drop in performance from $89\%$ to $81\%$.
The results in the fourth test (mixed-textures) show a lower performance for all the agents. DQN-multi has a success rate of $84\%$ and the DQN-single of $9\%$. The human pilots have a performance of $88\%$ and the AR-tracker of $82\%$.
The results of the fifth test (photo-realistic environments) show a generic drop (DQN-multi=$57\%$, DQN-single=$5\%$, Human=$81\%$, Random=$3\%$, AR-tracker=$84\%$). The overall performance on uniform textures, mixed textures and realistic worlds is $85\%$ for DQN-multi, $32\%$ for DQN-single, $88\%$ for human pilots, and $92\%$ for the AR-tracker. Finally, the results on the sixth test (real-world environment, mezzanine) showed an overall accuracy of $50\%$ on a total of 10 flights. We must point out that this condition was very challenging because of high variability in lighting and attitude instability.

%It is possible to further analyze the DQN-multi policy observing the utility distribution in different states (Figure~\ref{fig:fig_dqn_distribution}). When the drone is far from the marker the DQN penalizes the trigger action. However when the drone is over the marker the utility significantly increases triggering the vertical descent.

%The bar chart compares the performances of the DQN-single, DQN-multi, human subjects and random agent. The DQN-multi has the highest reported landing success rate with an overall accuracy of $89\%$. The score obtained by the agent trained on a single texture (DQN-signle) are significantly lower ($38\%$). The human performance is close to the DQN-multi ($86\%$). The performance on the different classes of textures shows that the DQN-multi obtained top performances in most of the environments. The DQN-single had good performances only on two textures: asphalt and grass. We verified using a two-sample t-test if the difference between DQN-multi and human pilots was statistically significant. We obtained a $t=2.37$ and $p<.05$ meaning that the difference is significant and the DQN outperformed humans.

\begin{figure}
    \centering
    \includegraphics[width=\textwidth]{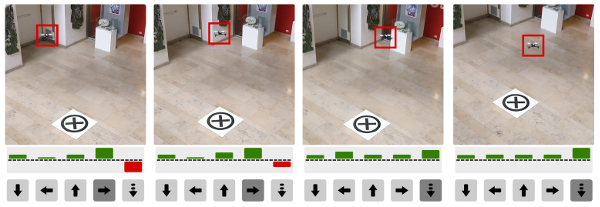}
    \caption{Snapshots representing vertical descent in the large hall environment. The bottom bar is the utility distribution of the actions. Descent has a negative utility (red bar) when the drone is not centred on the marker.}
    \label{fig:fig_dqn_distribution}
\end{figure}

\subsection{Second series of simulations} \label{ssec:second_simulation}
In the second series of simulations we trained and tested the DQNs specialized in the vertical descent. To encourage the descent during the $\epsilon$-greedy action selection we sampled the action from a non-uniform distribution where the descending action had a probability $\rho$ and the other $N$ actions a probability $\frac{1-\rho}{N}$. We used \textit{exploring-start} generating the UAV at different altitudes and ensuring a wider exploration of the state space. Instead of the standard buffer replay we used the partitioned buffer replay described in Section \ref{ssec:partitioned_buffer_replay}. We trained two networks, the former in a single texture condition (DQN-single) and the latter in multi-texture condition (DQN-multi).

\subsubsection{Methods}
The training environment was represented by a flat floor of size $100 \times 100$ m with the landmark positioned in the center.
The state-space in the vertical descent phase is significantly larger than in the marker detection and exploration is expensive. For this reason we reduced the number of textures used for the training, randomly sampling 20 textures from the 71. We can hypothesize that using the entire training set can lead to a better performance.
The action space available was represented by five actions: forward, backward, left, right, down. A single action was repeated for 2 seconds leading to an approximate shift of 1 meter due to a speed of 0.5 m/s. The descent action was performed at a lower speed of 0.25 m/s to reduce undesired vertical shifts.
The target and policy networks were synchronized every $C=30000$ frames. For the partitioned buffer replay we chose $\rho^{+}=0.25$, $\rho^{-}=0.25$, and $\rho^{\sim}=0.5$. A time limit of 80 seconds (40 steps) was used to stop the episode and start a new one. The drone was spawned with a random orientation inside a bounding box of size $3 \times 3 \times 20$ m at the beginning of the episode. This bounding box corresponds to the target area of the landmark detection phase described in Section \ref{ssec:first_simulation_methods}.
A positive reward of 1.0 was given only when the drone entered in a target-zone of size $1.5 \times 1.5 \times 1.5$ m, centered on the marker (Figure \ref{fig:fig_simulator_bounding_boxes_b}). If the drone descended above 1.5 meter outside the target-zone a negative reward of -1.0 was given. A cost of living of -0.01 was applied at each time step. 
The same hyper-parameters described in Section \ref{ssec:first_simulation_methods} were used to train the agent. In addition to the hardware mentioned in Section \ref{ssec:first_simulation_methods}, we also used a separate machine to collect preliminary experiences. This machine is a multi-core workstation with 32 GB of RAM and a GPU NVIDIA Tesla K-40. Before the training, the buffer replay was filled using a random policy with $10^{6}$ neutral experiences, $5 \times 10^{5}$ negative experiences and $6.2 \times 10^{4}$ positive experiences. We increased the number of positive experiences using horizontal/vertical mirroring and consecutive 90 degrees rotation on all the images stored in the positive partition. This form of data augmentation increased the total number of positive experiences to $5 \times 10^{5}$. To  test  the  performance  of  the  agents  we  measured the landing  success  rate  of  DQN-single, DQN-multi, human pilots, AR-tracker, and random agent in five tests. (i) In the first test the agents performed landing on 21 unseen uniform textures. (ii) The second test consisted in landing on uniform textures with a corrupted marker ( Figure~\ref{fig:fig_simulator_textures_sample}-i). (iii) In the third test 25 textures have been randomly sampled from the test set and mixed in a mosaic-like composition. (iv) In the fourth test landing has been accomplished in three photo-realistic environments: warehouse, disaster site, powerplant (Figure~\ref{fig:fig_simulator_textures_sample}-e/g). (v) In the fifth and last test the UAV had to land in four real-world indoor environments: laboratory, small hall, large hall, mezzanine (Figure~\ref{fig:fig_simulator_textures_sample}-a/d).
The performance of human pilots has been measured in all the simulated environments through discrete and a continuous controllers using the same procedure described in Section \ref{ssec:first_simulation_methods}.

%The networks were tested on the same 21 unseen textures used in the marker detection test. Performance of random agent and human pilots has been also collected. The human data has been obtained using a sample of 7 subjects. The subjects had to adjust the vertical position of the drone in order to move toward the marker and obtain a positive reward. The same procedure described in Section \ref{ssec:first_simulation_methods} has been used.

\subsubsection{Results}
\begin{figure}
\centering
\includegraphics[width=\textwidth]{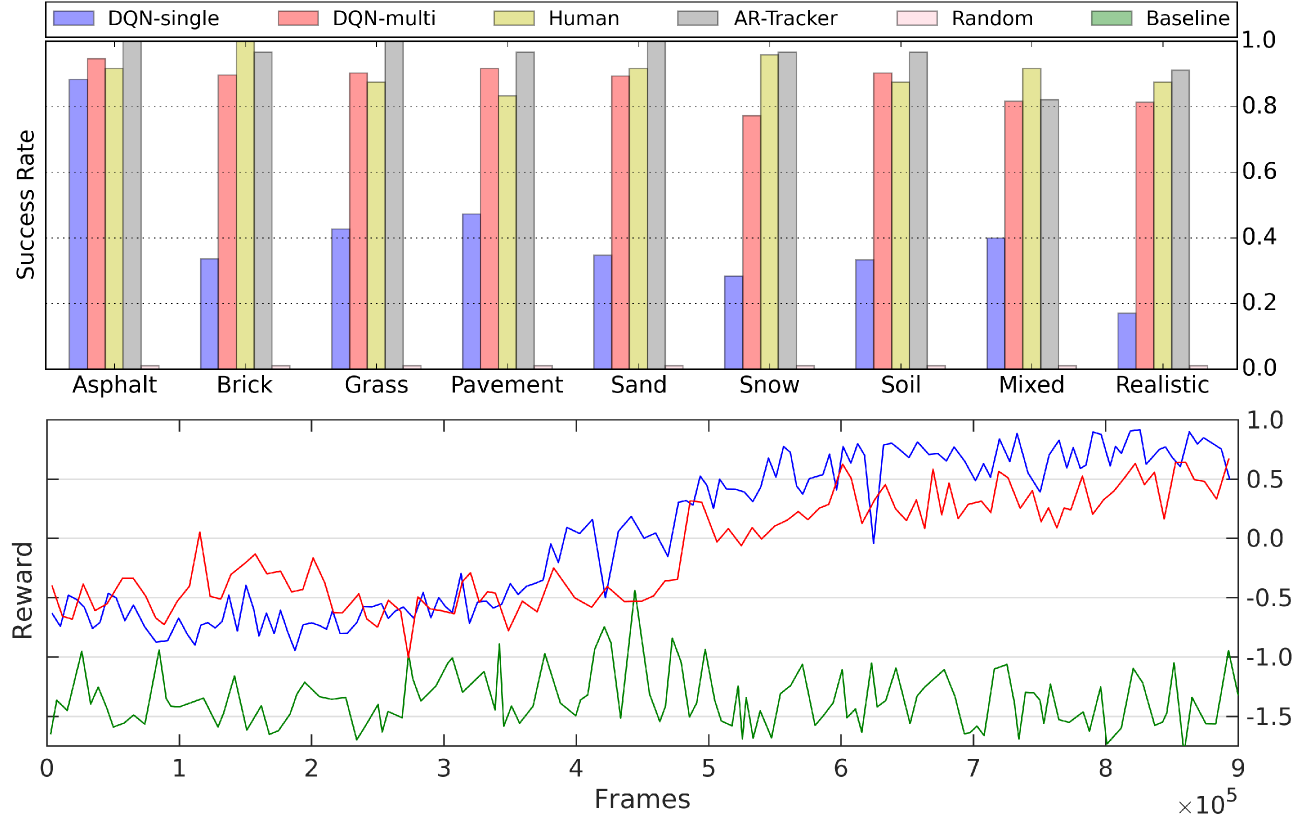}
\caption{Results of the second series of simulations. Top: descending success rate. Bottom: accumulated reward per episode for DQN-single (blue line), DQN-multi (red-line), and baseline (green-line).}
\label{fig:fig_results_second_series}
\end{figure}

the accumulated reward per episode showed in Figure~\ref{fig:fig_results_second_series} (bottom), increased stably in both DQN-single and DQN-multi. We reported also the baseline curve of a network trained on both detection and descent which did not learn to accomplish the task.The results of the test phase are summarized in Figure~\ref{fig:fig_results_second_series} (top). The bar chart compares the performances of the DQN-single, DQN-multi, human pilots, AR-tracker, and random agent. For human pilots we only report the score in the discrete control condition that is higher respect to the space-navigator condition ($+4\%$). The average score on the first test (uniform textures) is $89\%$ for DQN-multi, $44\%$ for DQN-single, $91\%$ for humans, and $98\%$ for the AR-tracker. Since both human pilots and DQNs used discrete steps to control the drone, it is possible to estimate the average number of steps required to accomplish landing. For human pilots the average number of steps is 23, whereas for the DQN-multi is 19, meaning that human pilots were slower. In the second test we compared the performances of the DQN-multi and AR-tracker on uniform textures with corrupted marker. The AR-tracker had a significant drop from $98\%$ to $0\%$ due to the failure of the underlying template-matching algorithm. The DQN-multi had a drop from $89\%$ to $51\%$ showing to be more robust to marker corruption. The third test (mixed textures) showed a general drop (DQN-multi= $82\%$, DQN-single=$40\%$, Human=$92\%$, Random=$1\%$, AR-tracker=$82\%$). In the fourth (realistic environments) have been observed a similar drop (DQN-multi= $81\%$, DQN-single=$17\%$, Human=$88\%$, Random=$1\%$, AR-tracker=$91\%$). The overall performances on uniform textures, mixed textures and realistic environments are $87\%$ for DQN-multi, $41\%$ for DQN-single, $91\%$ for human pilots, and $96\%$ for the AR-tracker. In the fifth and last test (real-world) the DQN-multi has been used to control the descending phase in 40 flights equally distributed in four environments (laboratory, small hall, large hall, mezzanine). The system obtained an overall success rate of $62\%$. Most of the missed landing have been caused by extreme light conditions (e.g. mutable natural light), and by flight instability (e.g. strong drift).
We can further analyze the DQN-multi policy observing the action-values distribution in different states (Figure~\ref{fig:fig_dqn_distribution}). When the drone is far from the marker the DQN penalizes the descent. However, when the drone is over the marker this utility significantly increases overcoming the others.

\section{Conclusions and discussion}\label{sec:conclusion}
%Discussion: This should explore the significance of the results of the work, not repeat them. Avoid extensive citations and discussion of published literature.
%Conclusion: The main conclusions of the study may be presented in a short Conclusions section, which may stand alone or form a subsection of a Discussion or Results and Discussion section.
In this work we used DRL to realize a system for the autonomous landing of a quadrotor on a static pad. The main modules of the system are two DQNs that control the UAV in two delicate phases: landmark detection and vertical descent. 
Using domain randomization we trained the DQNs with simple uniform textures and tested them in complex environments (both simulated and real). The overall performances are comparable with an AR-tracker algorithm and human pilots. In particular, the system is faster than humans in reaching the pad and is more robust to marker corruption compared to the AR-tracker. The most remarkable outcome is that the networks were able to generalize to real environments despite training performed on a limited subset of textures. In all the missed landing the flight was interrupted because of the expiration time. Not even once the drone landed outside of the pad. Most of the missed landing have been caused by extreme conditions (mutable lighting and strong drift), not modeled in the simulator. We hypothesize that the results can be further improved taking into account these factors during the training phase. In conclusion, the results obtained are promising, however further research is necessary in order to train stable policies that can effectively work in a wide range of real-world conditions.

%Despite training has been done on a limited subset of textures the networks were able to generalize on non-trivial realistic environments.
%We showed that the system can achieve super-human performances in the marker detection phase and almost-human performance in the vertical descent. Moreover we showed that we can train robust networks for navigation in large three-dimensional environments by training on multiple maps with random textures.

%Future work
%Future work should mainly focus on bridging the reality gap. The reality gap is the obstacles that makes it difficult to implement many robotic solutions in real world. This is especially true for DRL where a large number of episodes is necessary in order to obtain stable policies. Recent research worked on bridging this gap using domain transfer techniques. An example is domain randomization \cite{tobin2017domain}, a method for training models on simulated images that transfer to real images by randomizing rendering in the simulator. 

%Another method recently released is the CAD2RL \cite{sadeghi2016cad}. The CAD2RL is based on a deep network trained with Monte Carlo policy evaluation and it has been used to perform collision-free indoor flight in the real world while being trained entirely on 3D CAD models.

\subsection*{Acknowledgements}
We gratefully acknowledge the support of NVIDIA Corporation with the donation of the Tesla K40 GPU used for this research.

%%%%%%%%%%%%%%%%%%%%%%%%%%%%%%%%%%%%%%%%%%%%%%%%%%%%%%%%%%%%%%%%%%%%%%%%%%%%%%%%

\bibliographystyle{IEEEtran}      % basic style, author-year citations
\bibliography{mybibfile}          % name your BibTeX data base

\addtolength{\textheight}{-12cm}   % This command serves to balance the column lengths

\end{document}